\documentclass[10pt,twocolumn,letterpaper]{article}

\usepackage{wacv}              

\usepackage{graphicx}
\usepackage{amsmath}
\usepackage{amssymb}
\usepackage{booktabs}

\usepackage[pagebackref,breaklinks,colorlinks]{hyperref}

\usepackage[capitalize]{cleveref}
\crefname{section}{Sec.}{Secs.}
\Crefname{section}{Section}{Sections}
\Crefname{table}{Table}{Tables}
\crefname{table}{Tab.}{Tabs.}

\def\wacvPaperID{34} 
\def\confName{WACV}
\def\confYear{2025}

\begin{document}

\title{Adaptive Clustering for Efficient Phenotype Segmentation \\ of UAV Hyperspectral Data}

\author{Ciem Cornelissen, Sam Leroux, Pieter Simoens \\
IDLab, Department of Information Technology at Ghent University - imec\\
Technologiepark 126, B-9052 Ghent, Belgium \\
{\tt\small \{ciem.cornelissen, sam.leroux, pieter.simoens\}@ugent.be}}

\maketitle

\begin{abstract}
Unmanned Aerial Vehicles (UAVs) combined with Hyperspectral imaging (HSI) offer potential for environmental and agricultural applications by capturing detailed spectral information that enables the prediction of invisible features like biochemical leaf properties. However, the data-intensive nature of HSI poses challenges for remote devices, which have limited computational resources and storage. This paper introduces an Online Hyperspectral Simple Linear Iterative Clustering algorithm (OHSLIC) framework for real-time tree phenotype segmentation. OHSLIC reduces inherent noise and computational demands through adaptive incremental clustering and a lightweight neural network, which phenotypes trees using leaf contents such as chlorophyll, carotenoids, and anthocyanins. A hyperspectral dataset is created using a custom simulator that incorporates realistic leaf parameters, and light interactions. Results demonstrate that OHSLIC achieves superior regression accuracy and segmentation performance compared to pixel- or window-based methods while significantly reducing inference time. The method’s adaptive clustering enables dynamic trade-offs between computational efficiency and accuracy, paving the way for scalable edge-device deployment in HSI applications.
\end{abstract}

\section{Introduction}
\label{sec:intro}
Hyperspectral imaging (HSI) represents a step forward in data acquisition, capturing detailed spectral information across hundreds of narrow bands. Unlike conventional imaging systems, which are limited to a few broad spectral bands, HSI provides rich insights into the chemical and physical properties of an object \cite{hsiChemical}. This additional information enables predictions about features that are otherwise invisible, such as the molecular composition of plants or materials \cite{molecularComposition}. However, the data-intensive nature of HSI also presents significant challenges, particularly for remote observation devices like drones, satellites, or remote robots \cite{remotehsi}.

Remote devices have strict constraints on computational power, storage, and communication bandwidth. Storing the entire volume of hyperspectral data onboard is impractical, and transmitting it to centralized data centers for processing can be both slow and resource-intensive\cite{rs13050850}. Adaptive, efficient algorithms are essential for processing this data directly on remote devices. By reducing data footprints in an online matter, these algorithms make it feasible to store key information locally or transmit only the most relevant information to external storage or processing hub. As such this paper puts forward an efficient algorithm that makes it possible to process the output of a hyperspectral camera in real time, lowering the constraint on bandwidth and storage on remote devices. The model measures chlorophyll, carotenoid, and anthocyanin contents, producing a phenotype-segmented image which can be used for multiple insights. By tracking these metrics over time, it provides valuable data on their fluctuations, aiding in drought management and promoting forest health preservation. Additionally, it can also be employed for anomaly detection.

\begin{figure}[t]
  \centering
   \includegraphics[width=1\linewidth]{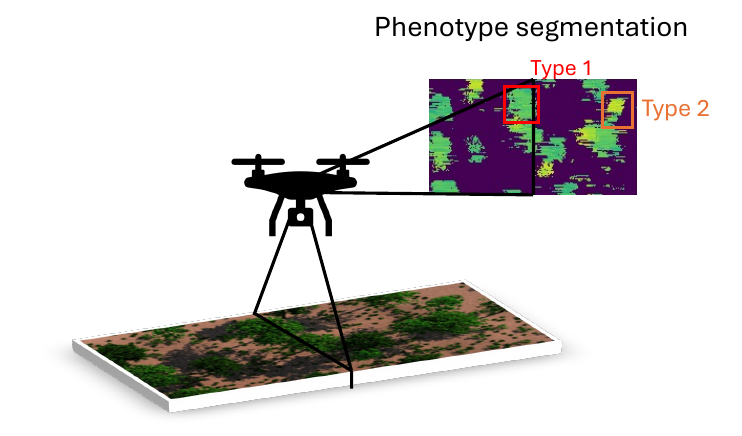}

   \caption{Illustration of the proposed real-time processing technique for the data of a hyperspectral push-broom camera for on-board phenotype segmentation of trees. }
   \label{fig:strartingPicture}
\end{figure}
HSI is particularly impactful in environmental monitoring and agricultural applications \cite{environmentalMonitoring}, where the spectral response of molecules provides critical insights into the chemical and structural properties of vegetation and landscapes \cite{propertiesvegetation}. For instance, hyperspectral data can detect specific molecular signatures, such as chlorophyll or water content, in plant leaves \cite{chlorophyllHSI}. This information can help identify stress factors like drought or disease early \cite{chlorophyllDroughtStress}, offering actionable insights for precision agriculture and conservation efforts. The ability to quantify these properties non-destructively and on a large scale sets HSI apart as a technology.

Adaptive algorithms further enhance the practicality of processing data on device in edge applications. By dynamically balancing resource constraints with computational demands, these algorithms optimize the trade-off between efficiency and accuracy. For example, edge-aware techniques can modulate their performance to the computational availability, reducing computational loads significantly without sacrificing the integrity of the analysis. This adaptability is crucial for deploying HSI systems on devices with limited power and processing capabilities.

In addition to advancements in UAVs and hyperspectral cameras, the growing computational power of modern systems has made high-fidelity simulations for HSI research possible \cite{riihiaho22}. Simulations enable the generation of synthetic data for scenarios where real-world data collection is infeasible or where labeled data is scarce. For example, by simulating the interaction of light with molecules in a realistic environment, one can create labeled hyperspectral datasets that predict molecular concentrations or other features of interest. These synthetic datasets can bridge the gap in applications where direct measurements are prohibitively expensive or technically challenging.

This work makes two key contributions: 
\begin{itemize}
    \item Simulation-driven Data Generation: an existing simulator is used and tweaked to create a high-fidelity simulation pipeline that generates labeled hyperspectral datasets by simulating light interactions with molecules in realistic environments. This approach bridges gaps in applications where real-world data is scarce, enabling robust training and validation of models under diverse scenarios. 
    \item Adaptive Edge Processing Algorithm \cref{fig:strartingPicture}: A novel algorithm gets proposed designed to process hyperspectral data efficiently on edge devices. By dynamically balancing resource constraints with computational demands, the algorithm reduces data footprints in real-time without sacrificing accuracy, making it suitable for deployment in resource-limited environments.
\end{itemize}

\section{Related Work}
\label{sec:Related Work}

\subsection{Hyperspectral Image Analysis}

Hyperspectral imaging (HSI) has become a critical tool in fields like remote sensing, agriculture, and environmental monitoring due to its ability to capture detailed spectral information across hundreds of narrow bands. Traditional approaches to HSI analysis often involve dimensionality reduction techniques, such as Principal Component Analysis (PCA) and Linear Discriminant Analysis (LDA), to manage the high dimensionality of the data while retaining relevant information \cite{Mantripragada_2022}. These methods, while effective in reducing computational complexity, often fail to account for noise, variability in real-world data, and applicability to a wide field of use cases \cite{HSI_comprerssionAlgos}. Having to compress the data in real-time makes it even harder\cite{rs13050850}. 

Deep learning has significantly advanced HSI analysis, with convolutional neural networks (CNNs) and recurrent neural networks (RNNs) being employed for tasks like classification and segmentation \cite{electronics12030488}. Methods such as 3D-CNNs leverage spatial-spectral information to improve segmentation performance \cite{10346933}. Despite their success, these approaches typically require large numbers of labeled data, which are often unavailable in HSI applications. Additionally, they are computationally intensive, making them unsuitable for real-time processing.

In the context of phenotyping, HSI has been applied to tasks such as crop classification, disease detection, and leaf chlorophyll estimation \cite{rs16183446}. Specifically in forestry, HSI has been used for species identification, health assessment, and estimating phenotypic traits like leaf area index and chlorophyll content \cite{forest_cholorphyll_est}. These studies highlight the potential of HSI for agricultural and forestry applications but are often limited by scalability issues and the computational inefficiency of their methods when applied to large datasets or real-time use cases. 

\subsection{Real-Time and Efficient Processing of HSI}

Real-time processing of HSI data remains a significant challenge due to the high dimensionality and volume of data acquired by hyperspectral sensors. Traditional approaches rely on simplifying assumptions or predefined thresholds to reduce computational overhead, such as band selection or down-sampling techniques \cite{GAO202031}. While these methods improve efficiency, they can lead to a loss of critical spectral information, impacting the accuracy of phenotyping tasks and reducing robustness to noise.

Incremental and adaptive algorithms have shown promise in real-time applications by processing data in smaller chunks, such as lines or patches, rather than the entire data point at once \cite{han2021dynamicneuralnetworkssurvey}. For instance, line-by-line processing methods have been explored for onboard fast anomaly detection in drone imaging \cite{diaz_guerra_horstrand_lopez_sarmiento_2019}. While these approaches enable real-time computation, they often lack mechanisms to handle noise effectively, as they may evaluate individual pixels without considering spatial or spectral correlations, leading to degraded performance in noisy or non-stationary conditions.

To enhance accuracy by reducing small amounts of noise, a clustering-based method has been proposed to group similar data points for collective processing \cite{waugh_allen_wightman_sims_beale_2018}. By grouping similar pixels one could reduce inherent noise and make algorithms more efficient through the collective processing of the pixels. Incremental or online clustering techniques, in particular, can dynamically adapt to new data while maintaining computational efficiency \cite{cohenaddad2019onlinekmeansclustering}. However, their application to HSI data, especially for phenotyping tasks, is still underexplored. Integrating clustering with robust classification models presents an opportunity to address both noise and efficiency challenges in real-time HSI analysis.

\begin{figure*}[th]
  \centering
   \includegraphics[width=1\linewidth]{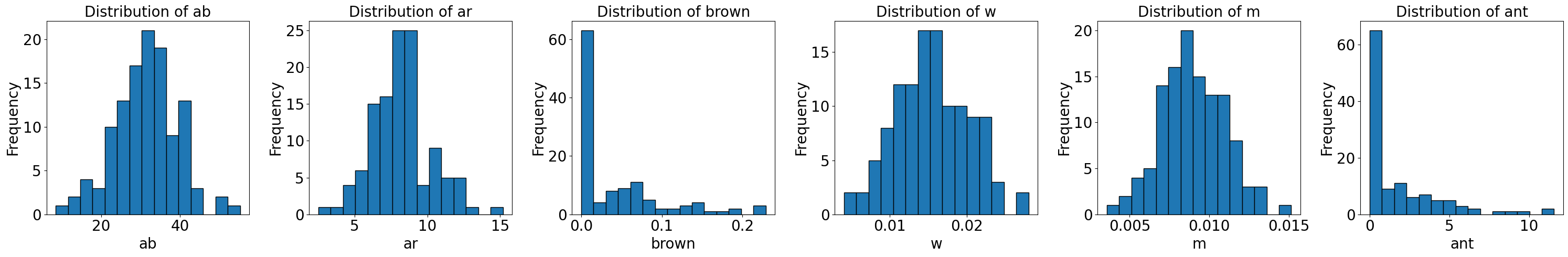}

   \caption{Distributions of the PROSPECT-D model parameters for the leaves in the generated data. This is to give an idea of the distribution of the labels in the dataset and of what values the different parameters can take. These are the values for chlorophyll, carotenoid, brown content, water thickness, dry matter, and anthocyanin respectively. }
   \label{fig:parameterDistributions}
\end{figure*}

\section{Methodology}
This section will start with an enhanced hyperspectral simulator \cite{riihiaho22}, which is modified to support parallel processing on large infrastructures. Next, the model is explained, integrating adaptive clustering with a neural network for classification and regression tasks. Finally, the training procedure gets described, emphasizing parameter tuning and optimization strategies.

\subsection{Hyperspectral Simulator and Data Generation}
The data used in the experimental section are generated through hyperspectral simulations. The simulator utilizes Blender to render the environment and employs the PROSPECT leaf model to generate realistic spectra and parameters. These spectra and parameters are provided to Blender for the final simulation. The simulator processes each wavelength separately; for each wavelength, Blender simulates light rays, modeling their interaction with the environment and determining whether they reach the camera. More detailed information about the simulation can be found in the HyperBlend project \cite{riihiaho22}.

For the generation of this dataset, fine-tuning of the software is necessary to enable it to run on various hardware platforms. Generating large datasets requires the capability to initiate multiple instances of the software, as a single generated image can take several hours. This is achieved by adjusting the software so that it's more suitable for parallelization on large infrastructures\footnote{This data is available on request}.

\begin{figure}[t]
  \centering
   \includegraphics[width=0.8\linewidth]{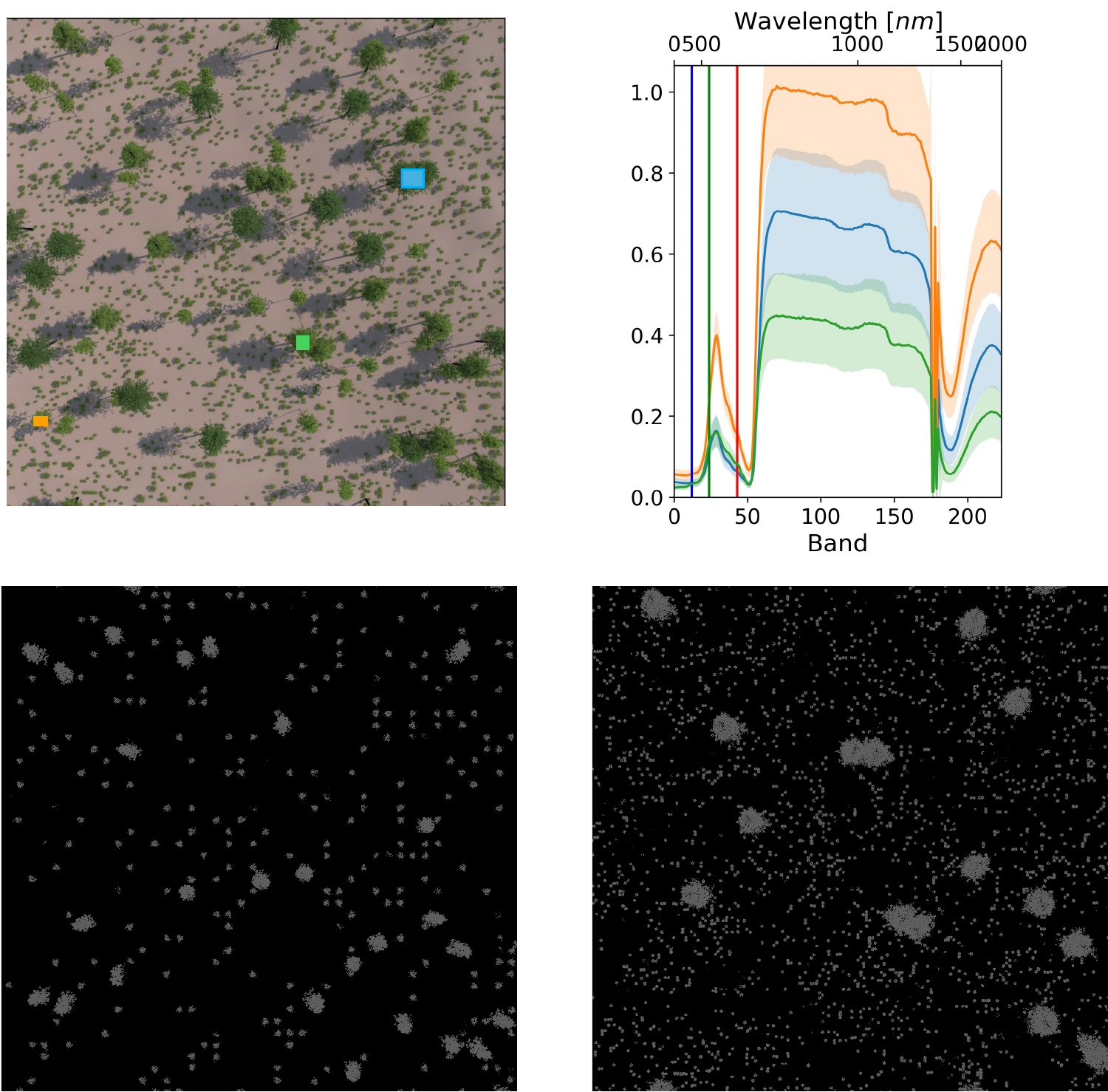}

   \caption{Example simulation data: fake RGB image on the top left, with spectra inside the highlighted squares on the top right. Best viewed in colour. The bottom to panel show the pixel labels for the trees. These contain values of 0 or 1 for if the tree is of certain class or not. }
   \label{fig:example simulation}
\end{figure}

To enhance the realism of the simulations, inspiration is drawn from \cite{horstrand_guerra_rodriguez_diaz_lopez_lopez_2019}, which describes a UAV platform equipped with a hyperspectral camera and provides details on on-board processing. The generated spectral bands in the simulations correspond to those detectable by the commercially available AFX10 and AFX17 hyperspectral cameras manufactured by SPECIM. These cameras are push-broom cameras, which means that the cameras record an HSI line per line. This is typical for hyperspectral cameras, as capturing full square images would result in being able to capture less spectral features. The simulated images are captured from the point of view of a drone positioned at an altitude of 200 meters, approximately ten times the height of the generated trees. This setup results in trees appearing approximately 50 pixels wide in the images. This is also visible from the results in  \cref{fig:example simulation}. The height is chosen to balance the number of trees one can capture with one sweep and the size of the trees. At this altitude, atmospheric effects would introduce noise into real-world data, which the simulator does not account for. However, there are programs available that correct for such noise \cite{shin2024robust, suomalainen2021direct}. 

Furthermore, atmospheric absorptions are considered for the incoming sunlight. As a result, the solar spectrum does not follow the energy curve of a theoretical blackbody \cite{spectrum_of_sun}, but instead contains absorption lines and bands from $H_2O$ and other molecules. This consideration makes it more challenging yet more realistic to use certain regions in the infrared spectrum, as these regions are almost fully absorbed by $H_2O$.

For calibration purposes, one simulation includes a white reference placed on the ground, which reflects $50\%$ of the incoming photons at each wavelength. This white reference is then used to calculate the reflectance values of all the leaves in all simulations. Normally, a dark reference is also required to account for the inherent noise the camera produces without incoming light. However, since these are simulations and the virtual camera does not have such issues, the reflectance values can be obtained using:
\begin{equation}
  R_{\lambda,i} = \frac{I_{\lambda,i}-D_{\lambda,i}}{W_{\lambda,i}-D_{\lambda,i}}
  \label{eq:reflectiveformula}
\end{equation}
In this equation $R_{\lambda,i}$ represents the reflectance at wavelength $\lambda$ for pixel $i$, $I_{\lambda,i}$ is the measured intensity at wavelength $\lambda$ for pixel $i$, $D_{\lambda,i}$ is the dark reference (zero in this case), and $W_{\lambda,i}$ is the white reference. Normally, the white reference is taken up close to calibrate individual pixels; however, here the white reference is observed from the air, making $W_{\lambda,i}$ independent of pixel $i$. Since the simulated camera does not exhibit variations between individual pixels, this approach is acceptable.

All simulations are performed with random values for various variables, such as leaf parameters, soil types, tree heights, branch lengths, and tree shapes. Adjusting these parameters introduces variability and enhances the realism of the simulations.

The PROSPECT model simulates leaf spectra, enabling the reconstruction of reflectance and transmittance values for wavelengths between 400 and 2500 nm \cite{PROSPECT}. The model has been updated over the years for increased accuracy and the inclusion of additional parameters. The version used in this simulation is the PROSPECT-D \cite{PROSPECT_codeReference} model. The leaf parameters available for this model are:

\begin{itemize}
    \item n (unitless), parameter of PROSPECT, this parameter resembles the structure of the leave. 
    \item ab ($\frac{\mu g}{cm^{2}}$), which is the chlorophyll $a + b$ content.
    \item ar ($\frac{\mu g}{cm^{2}}$), carotenoid content.
    \item brown (unitless), brown pigment.
    \item w (cm), equivalent water thickness. 
    \item m ($\frac{g}{cm^{2}}$), dry matter content
    \item ant ($\frac{\mu g}{cm^{2}}$), anthocyanin content. 
\end{itemize}

Chlorophyll is one of the most studied pigments in vegetation because it strongly affects the visible spectrum and plays a crucial role in photosynthesis. It serves as a good proxy for nitrogen content and is an indicator of environmental stress \cite{chlorophyll_droughtStress, chlorophyll_nitrogen}. Carotenoids are also important for photosynthesis, and anthocyanins help protect plants against UV radiation \cite{anth_carot}. The content distributions of the leaves are visualized in \cref{fig:parameterDistributions}.

As shown in the figure, the chlorophyll content of the generated leaves ranges from 10 to 50 ($\frac{\mu g}{cm^{2}}$), which is comparable to the chlorophyll content found in real leaves \cite{amount_of_chlorophyll}. The carotenoid content ranges from 4 to 14($\frac{\mu g}{cm^{2}}$), aligning with the carotenoid levels observed in actual foliage \cite{amount_of_carotenoid}. Anthocyanin contents are primarily close to 1~($\frac{\mu g}{cm^{2}}$), which is expected in healthy leaves. It is noted that anthocyanin content increases before leaves senesce and lose their green color \cite{anthocyanin}.

Each simulation randomly generates a forest on a rectangular area beneath the camera. The simulation provides pixel-level labels indicating the content of each pixel. Binary maps are generated for each pixel in the simulated area for leaf type, soil type, and bark type. If a pixel contains a leaf, its label is a vector of the seven parameters from the PROSPECT model. In \cref{fig:example simulation}, an example simulation is shown with the normalized spectra from Equation\ref{eq:reflectiveformula} displayed on the right. The colors of the spectra correspond to the colors of the squares in the synthetic RGB image on the left. From the different spectral distributions, it can be seen that the boxes are each centered on a different tree.

The full dataset consists of 180 simulated images, with each image taking around one hour to generate. Each image contains a randomly generated forest and tree randomly generated leaf types, with each tree having one type of leaf. The images have a resolution of $1024 \times 1024$ pixels, as most push-broom hyperspectral cameras capture lines that are 1024 pixels wide. The spectral data contain 224 bands: 112 bands from the AFX10 and 112 bands from the AFX17. These hyperspectral cameras are each capable of capturing 224 bands; however, in practice, the bands are often binned to reduce noise.

\subsection{Model Overview}
\begin{figure}[t]
  \centering
   \includegraphics[width=1\linewidth]{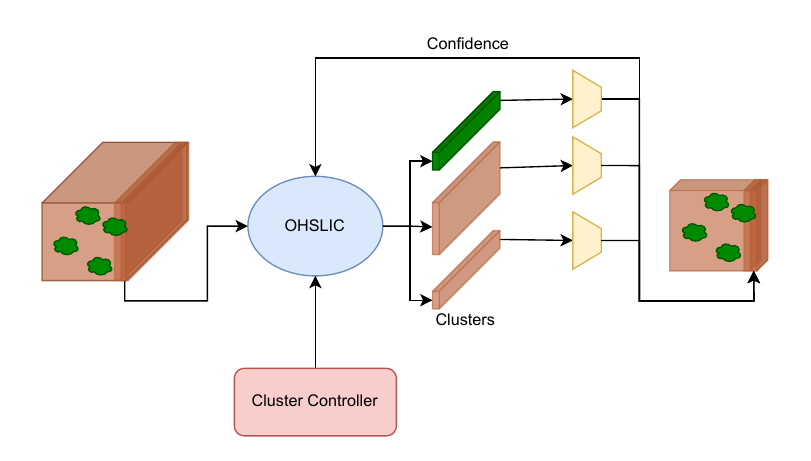}

   \caption{Illustrative figure of the model and the pipeline. OHSLIC takes a line as input and gives a set of clusters to the classifier network with multiple heads for the multiple predictions. The classifier network then gives its confidence back to OHSLIC and also outputs the labels for the different clusters. The number of clusters used in OHSLIC can be controlled by the clusters controller. }
   \label{fig:model}
\end{figure}

\begin{figure}[t]
  \centering
   \includegraphics[width=1\linewidth]{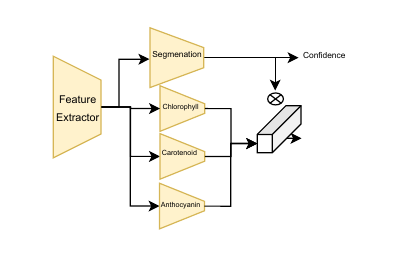}

   \caption{Illustrative figure of the Classifier model. The model is made of a 1D CNN feature extractor and 4 heads, which are MLPs of different sizes. The output of the model is a 1D array of length 3 for the three different parameters and the confidence of the segmenation.}
   \label{fig:classifier}
\end{figure}
The hyperspectral lines captured by push-broom cameras are inherently correlated in the direction of the camera’s movement. Sensor imperfections and environmental factors introduce noise, complicating the classification of individual pixels. To address these challenges, a new algorithm is proposed that builds on SLIC \cite{achanta2010slic} and online k-means clustering \cite{cohenaddad2019onlinekmeansclustering}, the Online Hyperspectral Simple Linear Iterative Clustering Algorithm (OHSLIC). This algorithm incrementally clusters pixels based on both spectral and spatial distances. Each pixel is assigned to the cluster with the minimal weighted cumulative distance (a combination of spectral and spatial distances). Cluster centroids are dynamically updated, with the frequency of updates adjustable based on computational constraints.

Once clustering is complete, the original pixel data and their cluster labels are passed to a classification network. This network determines whether each cluster represents a tree and, if so, predicts the properties of the tree’s leaves, such as chlorophyll content. To enhance robustness and mitigate pixel-level noise, the classification network processes the average spectral information for each cluster, rather than individual pixel data. This approach improves stability and accuracy in the predictions. The \textbf{C}lassifier then returns it's \text{C}onfidence of the prediction back to the clustering algorithm, giving the full algorithm OHSLIC-\textbf{C}-\textbf{C} shown in \cref{fig:model} and \cref{fig:classifier}. 

To identify the optimal classifier architecture, various configurations, and sizes were evaluated through a parameter sweep. For the feature extractor base, both a multi-layer perceptron and convolutional neural network are considered. For both these feature-extracting bases various depths are explored, with that also different sub-layers like dropout layers and maxpooling layers are investigated. Also for the different heads, different sizes are considered. Variations in output normalization and input normalizations are all considered at the same time. The selected design that gave the best results is a 1D CNN with four output heads as seen in \cref{fig:classifier}, with the feature extractor comprising three convolutional layers. One head is dedicated to segmentating pixels as either tree or background, while also providing confidence scores through applying a softmax function on the output of the classifier. The other three heads perform regression tasks, predicting specific tree parameters: chlorophyll, carotenoid, and anthocyanin levels. This architecture balances classification accuracy with precise multi-parameter regression. The final output of the network then contains 3 features for every pixel, which are 0 if it is not classified as a tree or the predictions for the parameters if the cluster is classified as a tree. 

The model’s performance is compared to a naive baseline approach, where all pixels are classified individually (PC), and a baseline where an average over a sliding window of 5 pixels is taken (APC). This comparison assesses OHSLIC’s impact on segmentation accuracy, regression quality, and computational efficiency.

The proposed model incorporates adaptive mechanisms to optimize computational cost and performance:
\begin{itemize}
    \item Confidence-Driven Cluster Refinement: The classification network monitors its confidence and dynamically splits clusters where significant mixing of tree and background pixels occurs.
    \item Computational Resource Adaptation: The cluster size and update frequency can be adjusted based on the platform’s computational capabilities or the required inference time. This adaptiveness ensures efficient operation in resource-constrained environments.
    \item Environmental Adaptation: Based on environmental factor, e.g. the height of the drone the clusters can be adjusted in size, to better suit the object that is being detected. 
\end{itemize}
Additionally, an adaptive area will be identified in the experimental section to enable flexible computation.

\subsection{Training Procedure}

The 1D CNN classifier, which can be seen in \cref{fig:classifier}, is trained end to end with it's 4 heads on average of pixels from a certain set of labels. The loss used during training is:
\begin{equation}
  Loss = Loss_{segmenation} + Loss_{regression}
  \label{eq:loss}
\end{equation}
Where $Loss_{segmenation}$ is the crossentropy between the output of the model and the labels of the data and $Loss_{regression}$ is Mean Squared Error(MSE) loss of each of the different head summed together. 
Training on the average of pixels drastically increased the performance of the network. This could be because taking averages over the pixel reduces some of that environmental noise making it easier for the network to make predictions because of more clear gradient flow because of a more clear pattern in the input data. 

For OHSLIC a sweep is done over different parameters of the algorithm. During this sweep things like spatial weight, spectral weight, and confidence threshold were adjusted to find the optimal set of parameters. The spatial weight and spectral weight are the weights used during the step where the distance between centroids and neighboring pixels is calculated. These weights are multiplied to difference in pixel distance and the difference in spectral distributions respectively. The optimal value of the spatial weight is around 40 and the optimal weight for the spectral weight is around 10. The confidence threshold is used to decide when a cluster should be split up and ranges from 0.5 to 1. Where 0.5 is that the algorithm is uncertain and 1 where the algorithm is certain about its prediction. The optimal value found for the confidence threshold is 0.7. 

For segmentation tasks, the Dice score \cite{dice} is a common metric, ranging from 0 (no overlap with ground truth) to 1 (perfect overlap). It is more forgiving of minor classification errors compared to metrics like IoU, making it more suitable here, where correctly segmenting trees is prioritized over perfect pixel classification. However, a challenge in this dataset is that smaller ground plants may share similar labels (e.g., chlorophyll content), complicating tree-only segmentation. To address this, only the Dice score for the background is evaluated, better reflecting the segmentation accuracy of the models.  

\section{Experiments}
\subsection{Performance OHSLIC-C}
To enable real-time processing, the algorithm is optimized and tested on a NVIDIA Jetson Nano, a compact device that can be used in UAV applications. Hyperspectral cameras typically achieve frame rates of 300 frames per second (fps). However, in practical scenarios, the fps is often reduced to around 50 fps to ensure sufficient photons are captured across all spectral bins, minimizing noise for lower-intensity wavelengths. This reduction imposes a hard processing limit of 20 ms per line.

\begin{figure}[t]
  \centering
   \includegraphics[width=1\linewidth]{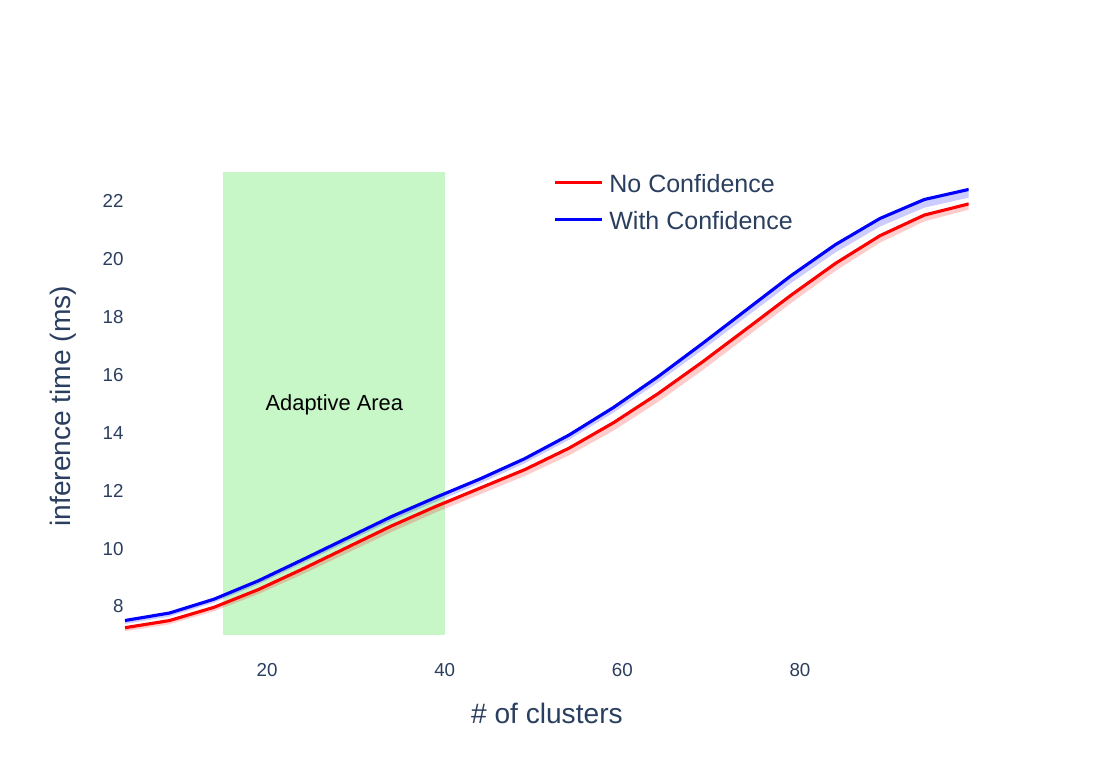}

   \caption{Inference time for processing a line in ms with respect to the number of clusters for OHSLIC-C with and without confidence. These times come from running the algorithm on a NVIDIA Jetson Nano.}
   \label{fig:numberClustersInferenceTime}
\end{figure}

As shown in \cref{fig:numberClustersInferenceTime}, the algorithm’s inference time is measured against the number of clusters. The figure also highlights an identified adaptive area where the cluster controller can modulate the number of clusters to maintain inference times between 8 ms and 12 ms, meeting real-time constraints. This adaptive range ensures flexibility in balancing computational efficiency and processing accuracy. This area is up to the user and can be made smaller or wider depending on one's needs. 

\begin{figure}[t]
  \centering
   \includegraphics[width=1\linewidth]{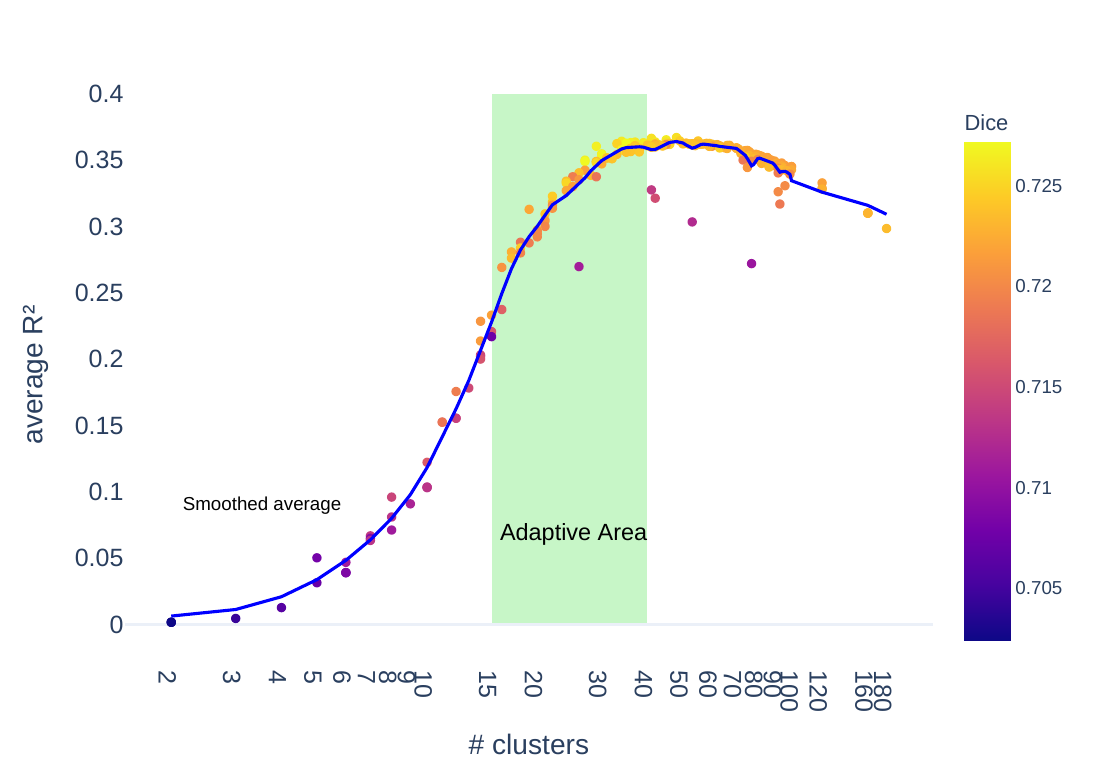}

   \caption{Average regression performance with respect to the number of clusters from OHSLIC and the Dice score.}
   \label{fig:regressionVSn_clusters}
\end{figure}

\cref{fig:regressionVSn_clusters} illustrates the average $R^2$ values for the regression of the three tree parameters as a function of the number of clusters. The results indicate that performance improves with an increasing number of clusters, peaking at the upper limit of the adaptive area before declining. This decline occurs because smaller cluster sizes amplify inherent environmental noise, reducing the accuracy of the regression. Both regression performance and segmentation accuracy peak at around 40 clusters. This means that the average size of the clusters is around 25 pixels, which is around the average size of the number of pixels a small tree takes up in a line. 

The figure also shows enhanced segmentation performance with finer clusters, which better capture intricate structures and patterns of the environment. Within the adaptive area, performance declines slightly as clustering granularity is reduced. To maintain optimal performance, the algorithm should operate near the upper limit of the adaptive area. However, in resource-constrained scenarios, the cluster controller can reduce the number of clusters, trading off some accuracy for computational efficiency.

\begin{figure}[t]
  \centering
   \includegraphics[width=0.7\linewidth]{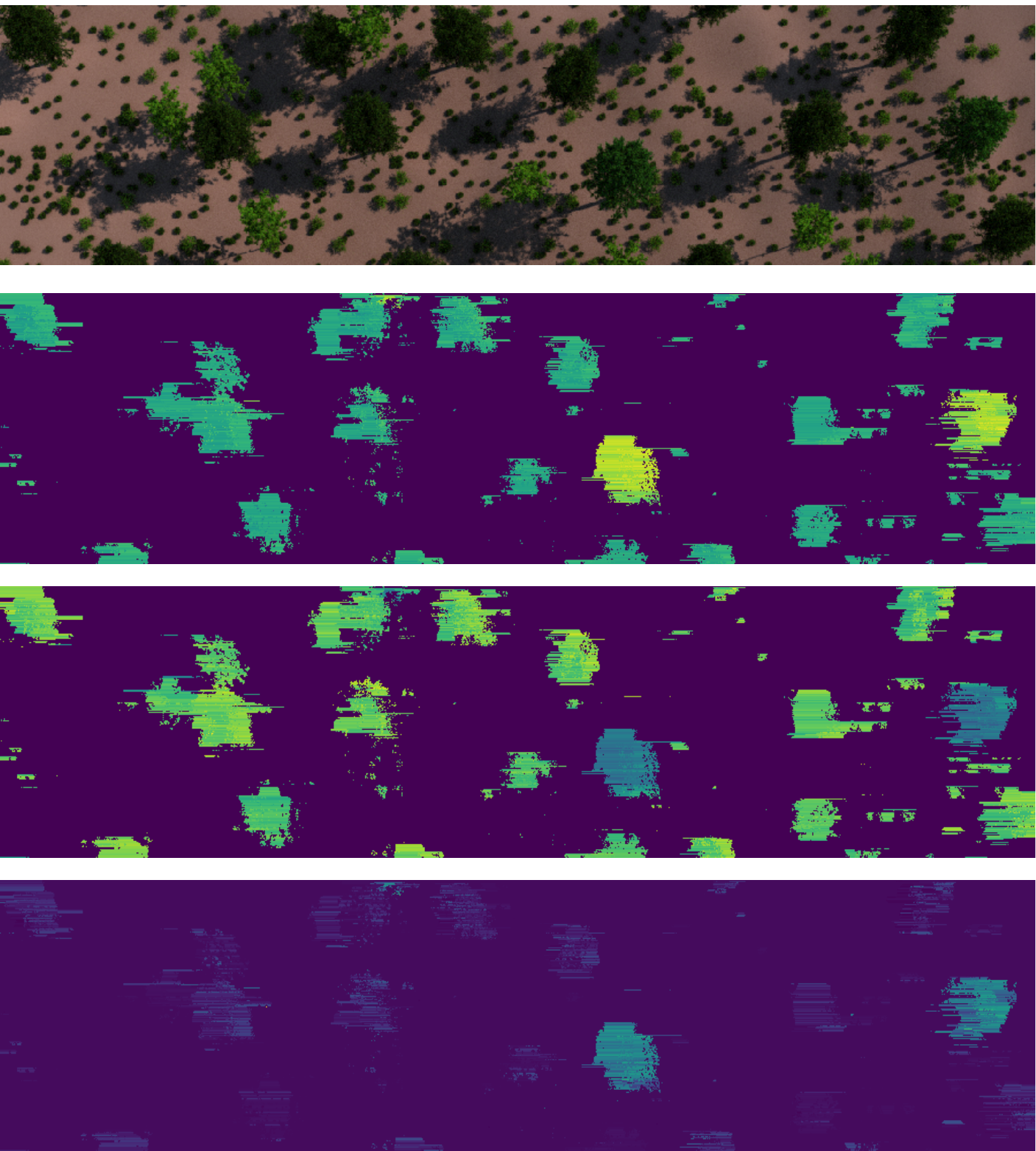}

   \caption{Example of the OHSLIC-C-C segmentation, with the top panel being a fake RGB image of the HSI. With the segmentation panels being chlorophyll, carotenoid, and anthocyanin respectively.}
   \label{fig:exampleSegmentation}
\end{figure}

\begin{figure}[t]
  \centering
   \includegraphics[width=1\linewidth]{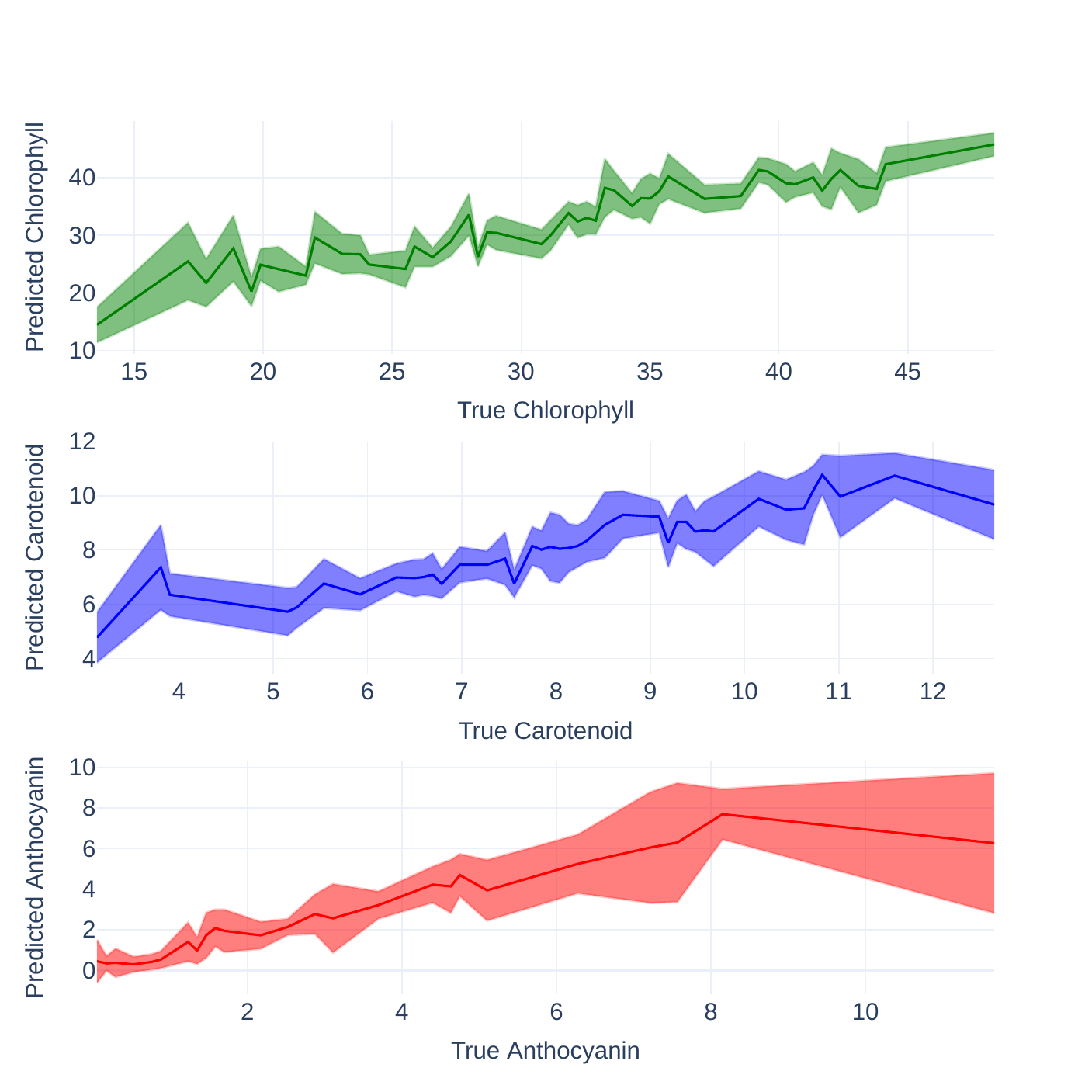}
    
    \caption{Regression predictions from the OHSLIC-C-C model vs the trued labels with there deviation.}
   
   \label{fig:example of regression predictions}
\end{figure}

\cref{fig:exampleSegmentation} shows an example of the segmentation capabilities of the network. The top panel displays an example of the data from the simulation, i.e. a fake rgb value from the hyperspectral image, while the lower panels illustrate the segmentation and regression results for the tree parameters. Regression of chlorophyll is the top panel of the segmentations. Here one can identify two trees that have a higher concentration of chlorophyll in their leaves. The cholorhyll content in those two trees is 38.4 ($\frac{\mu g}{cm^{2}}$) and the average predicted value is 38 ($\frac{\mu g}{cm^{2}}$). Below, the carotenoid and anthocyanin segmentations, respectively, further distinguish the two trees from the surrounding vegetation. For the 3 parameters, the two trees are easily distinguishable. Additionally, closer inspection of the carotenoid segmentation reveals two other distinct tree types, showcasing the network’s precision in segmentation and prediction of parameters. The overall regression accuracy for different values can be seen in \cref{fig:example of regression predictions}.

This tool can effectively distinguish outliers which might not be easily distinguishable by the human eye, mapping vegetation diversity within forests. Additionally, it measures chlorophyll, carotenoid, and anthocyanin contents, offering insights into plant health. By tracking these metrics over time, it provides valuable data on their fluctuations, aiding in drought management and promoting forest health preservation.

\cref{fig:example of regression predictions} shows the predictions for the three parameters, demonstrating that the model can accurately estimate content over a wide range of values. The figure makes it also clear that the predictions are still spread out for a certain set of contents, however, one must take into account that the predictions on, e.g. the edge of the tree, are also taken into account and that when looking more closely at the predictions in the centers of the trees the predictions become better and more closely distributed. It's thus more on the user of the data to select the right regions of the tree to see what the average prediction is. Or this could be automated in future works to detect the crowns of the segmented trees and to make these distributions smoother giving the predictions in the center of the circle higher weight then the predictions close to the edge of the circle, i.e. the tree. The model struggles more to predict values at the edges of the parameter distributions (\cref{fig:parameterDistributions}), likely due to fewer data points in these regions. To improve parameter prediction accuracy in future work, one could increase the number of data points, especially at the distribution extremes, ensuring better representation and coverage of the parameter space.

\subsection{Confidence}
\begin{figure}[t]
  \centering
   \includegraphics[width=1\linewidth]{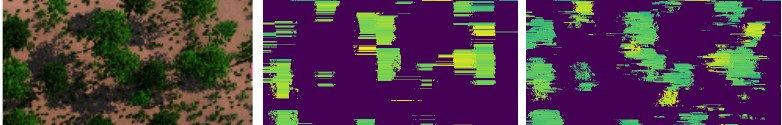}
    
    \caption{Example of how confidence changes the output of the model. On the left panel the fake RGB image of the dataset is visible. The middle panel is the normal segmentation without confidence and the right panel is the predictions with confidence. }
   
   \label{fig:comparisonconfidenceandnonconfidence}
\end{figure}

One drawback of using OHSLIC in a line-by-line approach is that the output can appear less structured, making it harder to visually distinguish individual trees. Additionally, focusing on a tree becomes challenging when it only partially appears in a single line. By incorporating confidence in predictions, OHSLIC-C-C improves the network’s ability to follow the natural structure of trees, resulting in clearer visual representations and enhanced segmentation and regression accuracy. This improvement comes with a minor increase in inference time as shown in \cref{fig:numberClustersInferenceTime}. 
\cref{fig:comparisonconfidenceandnonconfidence} illustrates this enhancement. This process works because mixed pixels-averages of both background and tree pixels are excluded during the classifier network’s training phase. Additionally, these mixed spectra fall between tree and background spectra, confusing the network and producing outputs closer to 0.5 after normalization. If this output falls below a predefined threshold, the cluster is flagged for splitting before the next line. This refinement improves segmentation accuracy and slightly enhances regression performance by ensuring that subsequent clusters better distinguish between tree and background pixels. It primarily enhances the alignment of the output with the input, providing a more accurate representation of the data.  
\subsection{Comparison}
To evaluate the impact of using OHSLIC, a comparison is made against the two baseline methods. \cref{tab:comparison} summarizes the segmentation accuracy, regression performance, and inference time for both approaches. Results demonstrate that OHSLIC-C-C, with confidence-based adjustments, significantly outperforms the baseline methods in predicting tree parameters. This highlights the effectiveness of using OHSLIC for clustering in simplifying the classification task, enabling the model to achieve better predictions while maintaining computational efficiency for real-time processing. OHSLIC-C-C makes it possible to process the line in real-time where as the baselines would have to reduce the fps of the camera significantly. 
\begin{table}
  \centering
  {\small{
  \begin{tabular}{@{}lcccc@{}}
    \toprule
    Method & Dice $\uparrow$ & $R^2$ $\uparrow$& inference time (ms) $\downarrow$& fps $\uparrow$\\
    \midrule
    PC & 0.718 & 0.091 & 86.2 $\pm$ 1.3 & 11.6 \\
    APC & \textbf{0.735} & 0.294 & 130.2 $\pm$ 0.5 & 7.7 \\
    OHSLIC-C & 0.724 & 0.362 & \textbf{12.2$\pm$ 0.1}& \textbf{82.0}\\
    OHSLIC-C-C & 0.727 & \textbf{0.367} & 12.4 $\pm$ 0.1& 80.6 \\
    \bottomrule
  \end{tabular}
  }}
  \caption{Performance of the different models. With $R^2$ the average $R^2$ of the three different parameters. The inference times are measured on a NVIDIA Jetson Nano. }
  \label{tab:comparison}
\end{table}





\section{Conclusion}
This study presents a novel approach to real-time hyperspectral data processing for tree phenotyping, leveraging adaptive clustering and robust classification. The OHSLIC framework achieves the best regression performance while maintaining computational efficiency, demonstrating its suitability for edge-device deployment in resource-constrained environments. A key contribution of this work is the development of a realistic hyperspectral dataset generated using a custom simulator, which incorporates physical and environmental factors to provide high-fidelity training data. Experimental results validate that clustering-based methods significantly outperform pixel-based alternatives, highlighting the advantages of noise reduction and grouped processing for accuracy and efficiency. Additionally, the framework’s adaptive capabilities allow users to balance accuracy and computational needs, offering flexibility for diverse applications. This work represents a step forward in scalable HSI analysis and provides a foundation for further exploration of adaptive algorithms for vegetation monitoring and beyond. Future research could address the challenges mentioned in the text or incorporate real-world data to validate the model’s performance under practical conditions.

\section{Acknowledge}
This work received financial support from the Flanders AI Research Program (FAIR). A special thanks to Kimmo Riihiaho for assistance in understanding the HyperBlend code.

{\small
\bibliographystyle{ieee_fullname}
\bibliography{egbib}
}

\end{document}